\documentclass[conference]{IEEEtran}
\IEEEoverridecommandlockouts
\usepackage{cite}
\usepackage{amsmath,amssymb,amsfonts}
\usepackage{algorithmic}
\usepackage{graphicx}
\usepackage{textcomp}
\usepackage{xcolor}
\def\BibTeX{{\rm B\kern-.05em{\sc i\kern-.025em b}\kern-.08em
    T\kern-.1667em\lower.7ex\hbox{E}\kern-.125emX}}

\usepackage{amsmath}
\usepackage{amssymb}

\usepackage{lipsum}

\usepackage{times}
\usepackage{epsfig}
\usepackage{graphicx}

\usepackage{multicol}

\usepackage{caption,subcaption}

\usepackage{gensymb}
\usepackage{float}
\usepackage{subcaption}
\usepackage{multirow}

\usepackage[utf8]{inputenc}
\usepackage[english]{babel}
\usepackage[normalem]{ulem}

\usepackage{soul}
\usepackage[norule,symbol,perpage]{footmisc}

\usepackage{booktabs}
\usepackage{booktabs}
\usepackage{graphicx}

\newcommand{\etal}{\textit{et al}. } 

\usepackage{soul}

\usepackage{soul}

\begin{document}

\title{Using CNNs to Identify \\ the
Origin of Finger Vein Sample Images}

\author{\IEEEauthorblockN{Babak Maser$^{\dagger}$\thanks{$^{\dagger}$ORCID iD: 0000-0002-1662-8324}}
\IEEEauthorblockA{\textit{ Multimedia Signal Processing \& Security Lab$^{\ddagger}$\thanks{$^{\ddagger}$http://www.wavelab.at/index.shtml}} \\
\textit{University of Salzburg}\\
5020 Salzburg, Austria \\
babak.maser@stud.sbg.ac.at}
\and
\IEEEauthorblockN{Andreas Uhl}
\IEEEauthorblockA{\textit{Multimedia Signal Processing \& Security Lab$^{\ddagger}$} \\
\textit{University of Salzburg}\\
5020 Salzburg, Austria\\
Uhl@cosy.sbg.ac.at}
}

\maketitle

\begin{abstract}
\textit{We study the finger vein (FV) sensor model identification task using a deep learning approach. So far, for this biometric modality, only correlation-based PRNU and texture descriptor-based methods have been applied. We employ five prominent CNN architectures covering a wide range of CNN family models, including VGG16, ResNet, and the Xception model. In addition, a novel architecture termed FV2021 is proposed in this work, which excels by its compactness and a low number of parameters to be trained. Original samples, as well as the region of interest data from eight publicly accessible FV datasets, are used in experimentation. An excellent sensor identification AUC-ROC score of 1.0 for patches of uncropped samples and 0.9997 for ROI samples have been achieved. The comparison with former methods shows that the CNN-based approach is superior and improved the results.}
\end{abstract}


\section{\textbf{Introduction}}
\label{sec:intro}

The authenticity and integrity of acquired \textit{finger vein} (FV) images play a significant role in the overall security of a finger-vein-based biometric system. In the advent of forgery techniques, it is important to link FV-images to their corresponding acquisition devices. A FV sample image not linked to a proper sensor of the recognition system would raise alarm and stop an eventual autentication process. Therefore, having reliable and trustful algorithms to achieve finger vein authenticity and integrity is vital.

\begin{figure}[hbt!]
    \centering
    \includegraphics[width=6cm,height=3cm]{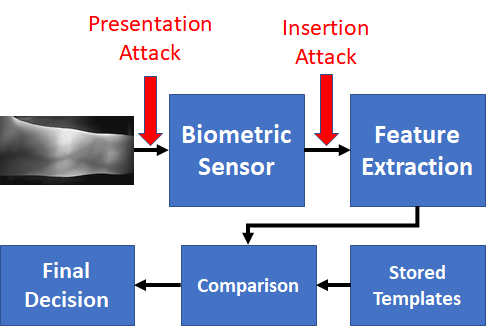}
    \caption{Points of insertion and presentation attack in a biometric system.}
    \label{fig:biometric_attack}
\end{figure}

Many biometric modalities (e.g. face, fingerprint, palm, finger vein images) are vulnerable to attacks.  Presentation attacks, which present spoof artifacts to the biometric sensor and insertion attacks, which bypass the sensor by inserting biometric samples into the transmission process between sensor and feature extraction module are the most important examples for attacking the user interface of the biometric system (see Figure \ref{fig:biometric_attack}). 

Sensor / camera identification in general can be achieved at different levels, camera model level, brand level, and device level. In biometric systems, we often intend to work on device level to actually uniquely identify the sensor instance having captured a certain sample. Still, sensor model identification is of interest as well \cite{maser2021identifying}: Securing a finger vein recognition system against insertion attacks in case the attacker does not know the employed sensor model and enabling device selective processing of the image data.

\vspace*{-1mm}

To identify the source device of an image, many algorithms have been proposed.
The most prominent way to deduce sensor information from images is to
exploit the image inherent Photo Response Non-Uniformity (PRNU). The PRNU relies on intrinsic characteristics of an image caused by different sensitivity of pixels to light due to inhomogeneity of silicon wafer and imperfection during fabrication of sensors.

Lukas \etal in \cite{lukas2006digital} and Fridrich in \cite{fridrich2009digital} propose to compute the residual image noise extracted from a subtraction of image and a denoised image version.  Linking between image and desired device is established by evaluating the similarity between the PRNU factor (also called the PRNU finger-print) and residual noise using NCC (Normalized Cross-Correlation). 

Another more recent way of source camera identification is based on deep learning specifically using CNN-based methods. Ahmed \etal  \cite{ahmed2019comparative} proposed a CNN model (three convolutional layers with a Softmax classifier) and compare the CNN-based result to a result based on PRNU, showing the PRNU-based approach to perform better than the proposed CNN approach. Baroffio \etal  \cite{baroffio2016camera} obtained good accuracy with a three convolutional layer CNN model on a larger dataset.
Tuma \etal \cite{tuama2016camera} obtained good results as well
looking into different CNN models, among them AlexNet, GoogleNet, and
an architecture proposed in their work.
Bondi \etal \cite{bondi2016first} propose a CNN model with four layers of convolution along with an SVM classifier instead of a fully connected layer for classification. Note that so far, CNN-based techniques have been proven to be successful in device model identification, while PRNU-based techniques are able to support both
model as well as device level identification.

Moving to the biometric domain, Bartlow \etal \cite{5204312} investigated on identifying sensors from fingerprint images using a PRNU-based technique. They examined effect and influence of sensor identification even when one only has access to a limited number of samples. Focusing on the iris subdomain, Kauba \etal \cite{Kauba18c} used   PRNU- and image texture-based methods. For the texture-based method, the authors extracted texture descriptors by applying DSIFT, DMD, and LBP. The extracted features are represented by Fisher Encoding  and discriminated by SVM. Banerjee \etal \cite{banerjee2017image}  evaluate the applicability of different PRNU estimation schemes to deduce sensor information from NIR iris images. Moving from classical  approaches to deep learning in iris sensor identification, Marra \etal  \cite{marra2018deep} proposed a three layer CNN architecture with a Softmax layer at the end.
 
Shifting from the iris to the FV subdomain, Maser \etal  \cite{maser2019prnu,maser2019prnuOAGM} and Söllinger \etal  \cite{sollinger2019prnu} applied PRNU-based sensor identification methods on finger vein datasets. 
Maser \etal proposed a texture-based sensor model identification approach \cite{maser2021identifying}, the features have been extracted by applying several classic and statistical properties of an image such as Histogram, Wavelet variance, Entropy, and LBP, etc.\ SVM is applied to discriminate the sample image origins.  In this work, we use their results in comparison to our research results. So far, deep learning based techniques have not been used to identify the origin of vascular sample data at all. 

In this work, we focus on identifying the origin of finger vein sample images using a CNN-based approach. Besides the evaluation of existing CNN models, we also propose a custom one and show its beneficial properties.

This work is structured as follows: In section \ref{sec:cnn_models} we describe the six used CNN models. Section \ref{sec:datast_Exp_design} discusses the properties of the finger vein sample datasets as considered and the setup of the conducted experiments. Next, we discuss and analyze the experimental results in Section \ref{sec:result}, and finally, we end this manuscript with a conclusion in Section \ref{sec:conclution}.

\section{\textbf{CNN models' structure}}
\label{sec:cnn_models}
In this section we discuss briefly five state-of-the-art CNN models, and introduce further a novel CNN model adapted for the target application.  
To select the most appropriate CNN models our attention is on examining a full range of prominent CNN models with varying properties, from a simple variance of AlexNet to more complex architectures like the Xception model which hopefully gives us a deep understanding which type of model is suitable to learn the patterns of finger vein samples. We also study the complexity of introduced models in Table \ref{tbl:number-of-params-in-cnn-model}.

\subsection{\textbf{Marra and Bonidi Models}}
\label{ssec:bondi_marra_models}
These two models are simple stacked networks (AlexNet family networks, both have been used in camera / sensor identification before). 

(i) \textbf{Bondi Model:} The CNN model proposed by Bondi \etal \cite{bondi2016first} is a stack of convolutional layers which end with a fully-connected layer. We adopted the model to preserve the given specification as good as possible. However, the only changes we made in their proposed model is that the SVM classifier is replaced by a Softmax layer. The detailed structure of the Bondi model is given in the above mention paper. Referring to Table \ref{tbl:number-of-params-in-cnn-model} Bondi model is a relatively light stacked network.

(ii) \textbf{Marra Model:}  Marra \etal proposed a network \cite{marra2018deep} which is an AlexNet variant, however, the number of layers has been reduced as compared to its predecessor. Due to having 2 fully-connection layers with 1024 and 2048 neurons, the number of trainable parameters is significantly increased (Table \ref{tbl:number-of-params-in-cnn-model}) which causes high complexity of the model. 

In the above section, we discussed two relatively shallow CNN networks (Bondi and Mara). Goodfellow \etal \cite{goodfellow2013multi} showed that the increasing depth of a network layer led to better performance. In other words, incrementing the number of layers in a network leads to gain enriched feature maps.  Thus, due to the mentioned fact, we wanted to know how various deeper networks will perform on our FV databases. 

\subsection{\textbf{VGG16 Model}}
\label{ssec:vgg16_model}
We employ the VGG16 model which has been introduced in \cite{simonyan2014very}. VGG16 is an example of a deep network, and, as a result, this model showed improvements in performance with respect to its predecessor models. Furthermore, VGG16 is designed to enrich the feature maps by expanding layers to have a deeper network compared to simple convolutional layers like described in Section \ref{ssec:bondi_marra_models}. 
We adopted ConvNet Configuration type B of the VGG16 network which is represented in Table 1 of \cite{simonyan2014very} with a slight modification in the input layer, and and adapted number of classes in $Softmax$. Also, we reduced the number of Convolutional layers with 512 channels from 4 to 2.

Even though deeper networks have advantages w.r.t more shallow networks, higher network depth may leed to another problem called \textit{degradation}. To avoid this potential problem, Residual networks have been introduced.   

\subsection{\textbf{50-Layer ResNet Model (ResNet50)}} \label{ssec:resnet_1_model}
The 50-layers Residual network exploits the concepts of deep residual learning. One problem of a deep network is \textit{degradation}, i.e. when a deep network starts converging, accuracy gets often saturated. Having accuracy saturation in a network implies that the model does not optimize well. In addition, a deep network leads to higher training errors. He \etal addressed these problems by introducing a deep residual network (a.k.a ResNet), therefore, we select the Residual network proposed by He \etal in \cite{he2016deep} as a further candidate. In summary, (i) the deep Residual network is easily optimized (training error decrease) as compared to its counterpart stacked networks, and, (ii) gains better accuracy while increasing the network depth. 
Further, (iii) the complexity of a Residual network is low compared to plain stacked CNN networks (referring to Table \ref{tbl:number-of-params-in-cnn-model}), e.g. A ResNet having 152 layers has less parameters than e.g. a VGG model which has been discussed in Section \ref{ssec:vgg16_model}. Details of the ResNet architecture is given in \cite{he2016deep}.

\subsection{\textbf{Xception Model}} \label{ssec:xception_model}
We selected a variation of the Inception model called Xception that is proposed by Chollet \cite{chollet2017xception}. The Xception model is claimed to be capable of learning with fewer parameters. The philosophy behind this architecture is to decouple the mapping of cross-channel correlations and spatial correlations in the feature maps of CNNs. To achieve decoupling, the depth-wise separable convolution is applied, which works as follows: A spatial convolution is executed independently over each channel of an input, then a point-wise convolution  ($1\times1$) is applied sequentially. The output of channels is projected by depth-wise convolution onto a new channel space.
It is important to mention that Xception applies a nonlinearity mapping after each operation in the depth-wise separable convolution process.
In summary, the Xception model is a linear stack of depth-wise separable convolution layers with a residual connection. The details of the Xception architecture is given in \cite{chollet2017xception}.

\subsection{\textbf{6-layer CNN Model (FV2021)}} 
\label{ssec:resnet_2_model}
\begin{figure}[hbt!]
    \centering
    \includegraphics[width=6cm,height=8cm]{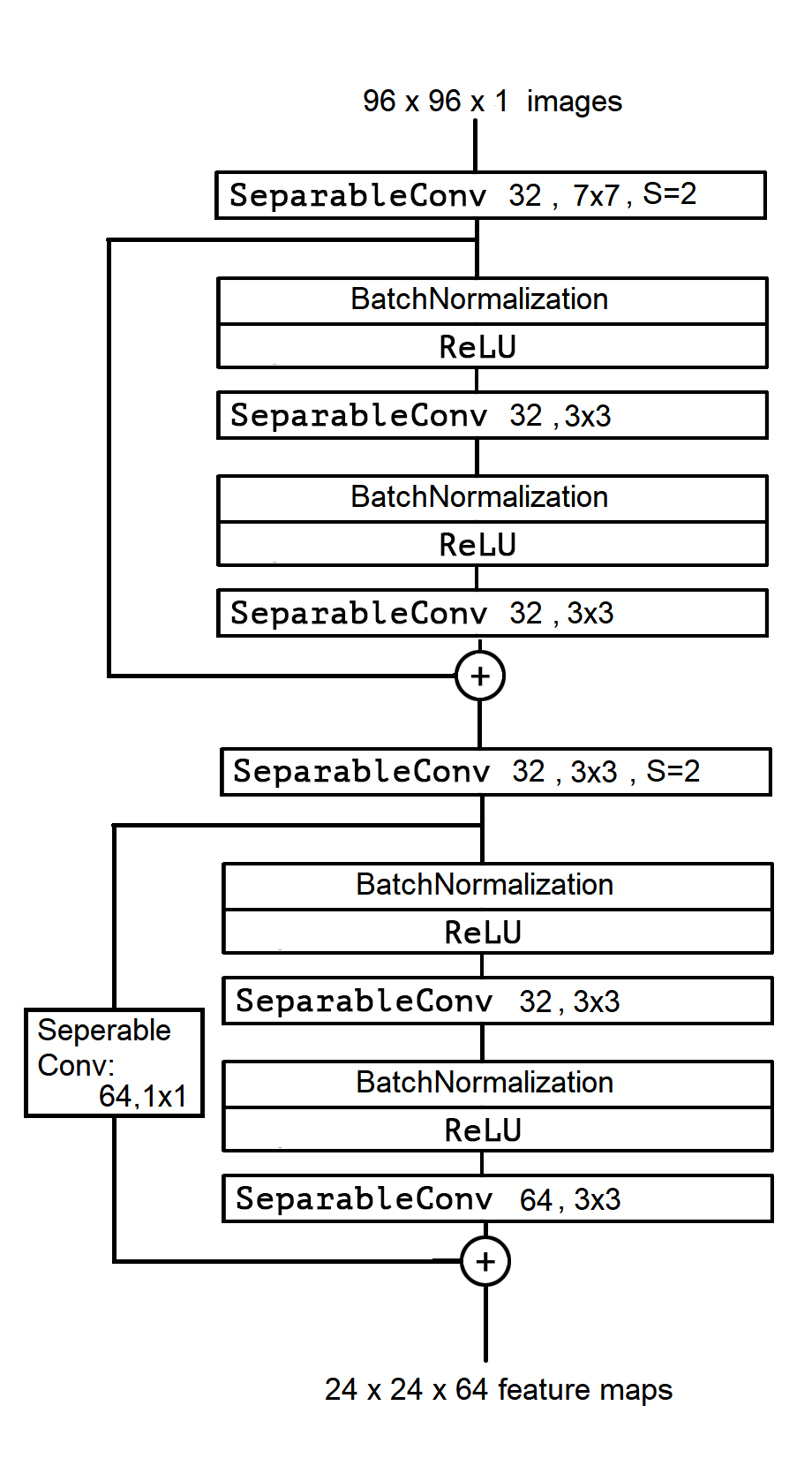}
    \caption{Proposed model: FV2021 CNN Architecture, a fully connected layer can be added optionally. }
    \label{fig:macro_resnet_model}
\end{figure}

To propose a novel network that has the advantage of being small and also exploits the advantage of the most prominent CNN models, we could think of many architectures, and most might also work. However, we propose a small model (eventually also well suited for a mobile device) and aim to achieve the same accuracy as the large CNN models. Thus, one of the advantages of the FV2021 model is having the lowest complexity (Table \ref{tbl:number-of-params-in-cnn-model}).  
We exploit the advantage of Separable Convolution (SC, as used in the Xception net) instead of the classic convolution layer. As explained before, separable convolution performs a depth-wise spatial convolution (which acts on each input channel separately) followed by a point-wise convolution that mixes the resulting output channels.
Thus, in developing FV2021, we took the advantage of cross-channel correlations as well as spatial correlations. Therefore, we applied and exploited the utilization of small receptive as well as $1 \times 1$ convolution filters, which can be seen as a linear transformation of the input channels.
The network architecture is composed of two sequential blocks, the first block has a skip connection, but the second block has a residual connection (a connection with a convolution operator). To reduce the computational complexity in the first layer, the number of filters are reduced to 32, and kernel size is $7\times7$ with strides of 2, each convolution layer is followed by a batch normalization and a nonlinearity unit (ReLU). A complete scheme of the proposed architecture is given in Figure 
\ref{fig:macro_resnet_model}, also with parameters given in each convolution block as follow: Separable Conv. $<$number of filters$>$, $<$receptive field size $>$, $<$s=strides$>$.

\vspace*{-1mm}
\vspace*{-1mm}

\subsection{\textbf{Complexity of CNN Architectures}}
\label{ssec:xception-or-fv2021}

Complexity and memory consumption are vital criteria to select an algorithm in general. Besides, in particular when selecting CNN models, the performance, resource consumption, and complexity of the model can be considered as Achilles heel for practical applications. One way to estimate the complexity and resource consumption in a CNN is to calculate the number of trainable parameters which are being used by a CNN architecture. We show the number of total and trainable parameters for each of discussed CNNs architecture in Table \ref{tbl:number-of-params-in-cnn-model}. In addistion, the last column shows the number of weighted layes. Please, note that in each model the last fully-connected layer (FC) includes the \textit{Softmax}.  

\begin{table}[!htbp]
\centering
\renewcommand\arraystretch{1.2}
\resizebox{0.4\textwidth}{!}{%
\begin{tabular}{l||rrr}
\multicolumn{1}{c||}{CNN Model} &
  \multicolumn{1}{c}{\begin{tabular}[c]{@{}c@{}}Total \\ params\end{tabular}} &
  \multicolumn{1}{c}{\begin{tabular}[c]{@{}c@{}}Trainable\\  params\end{tabular}} &
  \begin{tabular}[c]{@{}c@{}}Number \\ of  Layers\end{tabular} \\
  \hline \hline
Bondi    & 2,681,368  & 2,681,304  &  4 Conv + 2 FC \\
Marra    & 65,563,720 & 65,563,720 &  3 Conv + 2 FC \\
VGG16    & 55,097,288 & 55,077,064 &  8 Conv + 3 FC \\
ResNet50 & 23,597,832 & 23,544,712 & 50 Conv + 1 FC   \\
Xception & 20,877,296 & 20,822,768 & 36 Conv + 1 FC      \\
FV2021   & 314,632    & 314,376    &  6 Conv + 1 FC       \\
\end{tabular}
}
\caption{\small Number of total, trainable parameters and weighted layers in CNN models}
\label{tbl:number-of-params-in-cnn-model}
\end{table}

Table \ref{tbl:number-of-params-in-cnn-model} reveals that FV2021 has the minimum number of trainable parameters while other models have an enormous number of parameters, ranging from 2.5 to 65 millions.  Thus, the proposed CNN has the lowest complexity in comparison to the other discussed models in this section. We will discuss the sensor identification performance of the CNN architectures in Section \ref{sec:result}.



\section{\textbf{ Finger Vein Sample Data \& Experimental Design}}
\label{sec:datast_Exp_design}
We consider eight different finger vein databases (acquired with distinct prototype near infrared sensing devices) that are well known, in addition they are accessible publicly. In this work we took 120 samples from each database.
The databases are as follow:
\begin{enumerate}
    \item SDUMLA-HMT,\item HKPU-FV,\item IDIAP,\item MMCBNU\_6000 (MMCBNU),\item PLUS-FV3-Laser-Palmar (Palmar),\item FV-USM, \item THU-FVFDT,\item UTFVP.
\end{enumerate} 

Information on the size of the original samples and how the samples have been withdrawn from the datasets are given in \cite{sollinger2019prnu,maser2021identifying}.

\subsection{\textbf{Finger Vein Region of Interest(ROI)}}
\label{ssec:roi}

In finger vein recognition features are typically not extracted from a raw sample images but from a region-of-interest that is the portion of an image containing only finger vein texture. In addition, an insertion attack can also be mounted using ROI samples (in case the sensor does not deliver a raw sample to the recognition module but ROI data instead). Thus, we produced cropped image samples (ROI datasets) out of the original samples to be able to test our approach on these data as well. To produce ROI datasets we follow the same approach as it is proposed by Maser \etal in \cite{maser2021identifying}.

\begin{figure}
    \centering
    \begin{subfigure}[t]{.35\textwidth}
    \vspace{2pt}
    \centering
        \raisebox{-\height}{\includegraphics[width=0.45\textwidth]{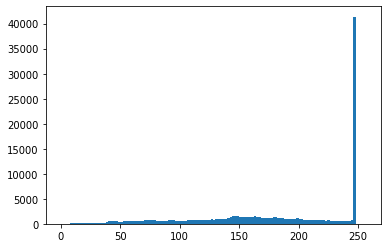}}%
        \hspace{1pt}
        \vspace{2pt}
        \raisebox{-\height}{\includegraphics[width=3.0cm,height=1.7cm]{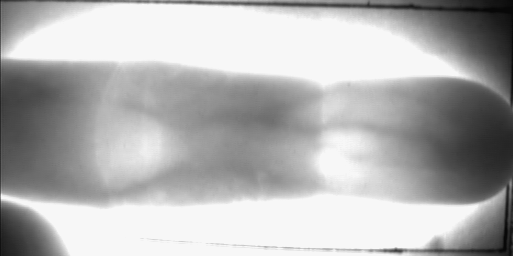}}%
        \hspace{1pt}
        \raisebox{-\height}{\includegraphics[width=0.45\textwidth]{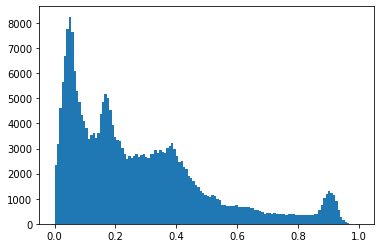}}%
        \vspace{.1ex}
        \raisebox{-\height}{\includegraphics[width=3.0cm,height=1.7cm]{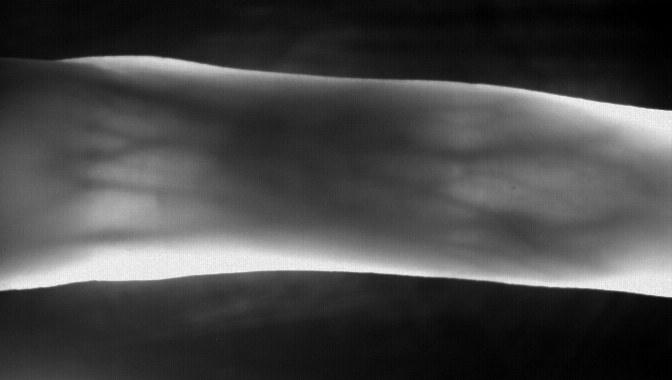}}%
        \hspace{1pt}
        \raisebox{-\height}{\includegraphics[width=0.45\textwidth]{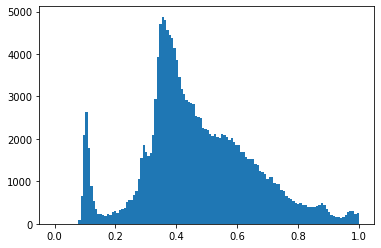}}%
        \hspace{1pt}
        \vspace{.5ex}
        \raisebox{-\height}{\includegraphics[width=3.0cm,height=1.7cm ]{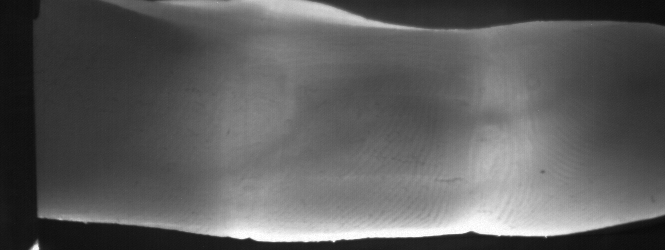}}
    \end{subfigure}
    \caption{FV Image samples and corresponding histograms of original sample images, from top: (a)\hspace{1pt}HKPU\_FV dataset, (b)\hspace{1pt} UTFVP dataset,(c)\hspace{1pt} SDUMLA dataset, and, (d)\hspace{1pt}IDIAP dataset.}
    \label{fig:histogram_datasamples}
	
\end{figure}

The original samples, as shown in Fig. \ref{fig:histogram_datasamples}, can be discriminated easily: Besides the differences in size (which can be adjusted by an attacker of course), the sample images can be probably distinguished by the extent and luminance of background. To illustrate this, we display the images' histograms beside each example in Figure \ref{fig:histogram_datasamples}, and those histograms
clearly exhibit a very different structure. Thus, we have learned that even texture descriptors
have an easy job to identify the origin of the respective original sample images. This is not necessarily the case for ROI data.

\begin{figure}[h!]
    \centering
 
    \includegraphics[width=0.49\linewidth]{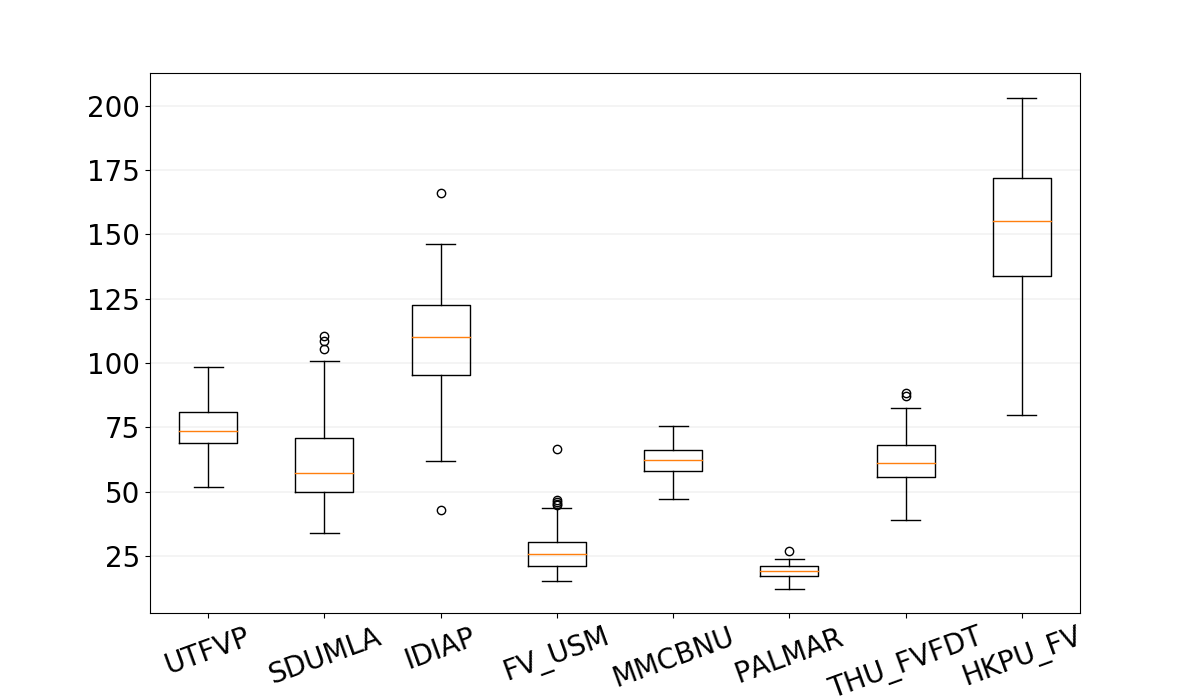}
    \includegraphics[width=0.49\linewidth]{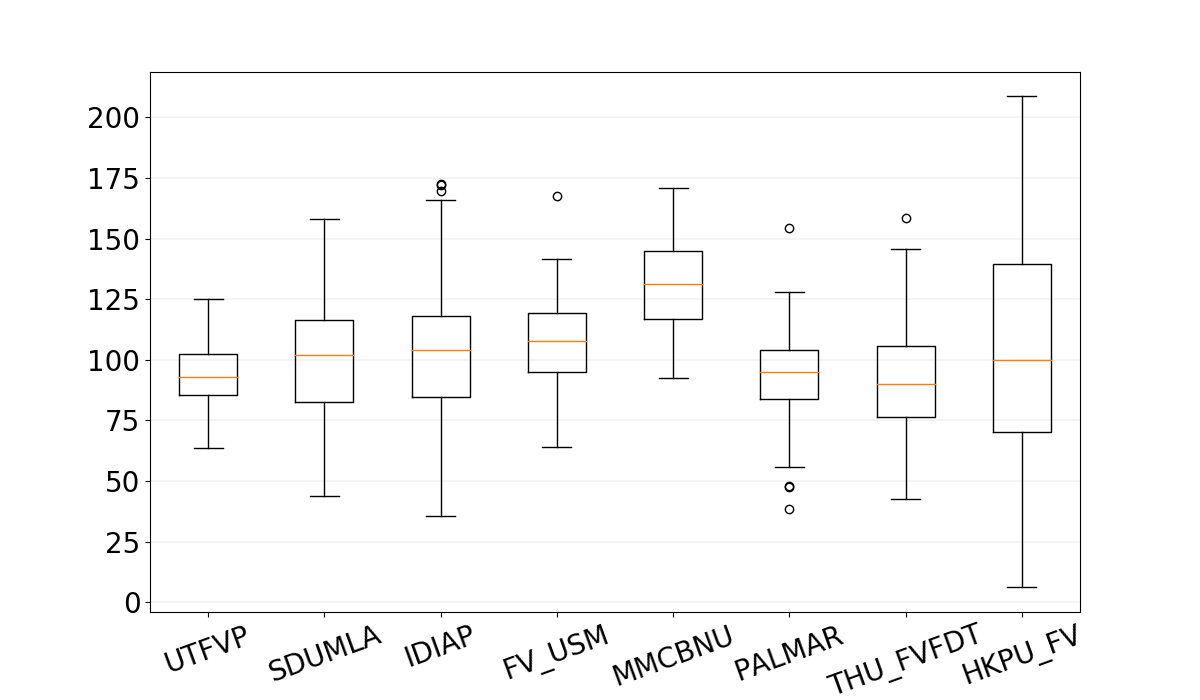}
    \caption{Luminance distribution of original and ROI images across all datasets, respectively.}
   
    
    \label{fig: luminance_uncropped}
\end{figure}

\begin{figure}[h!]
    \centering
    \includegraphics[width=0.49\linewidth]{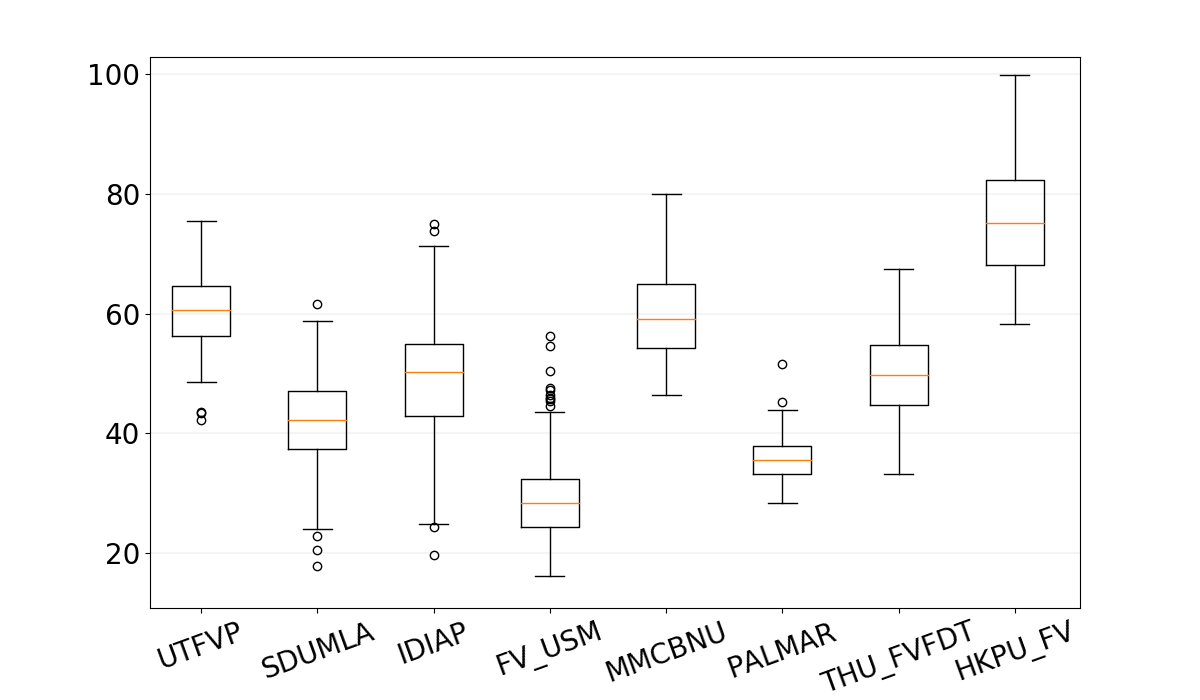}
    \includegraphics[width=0.49\linewidth]{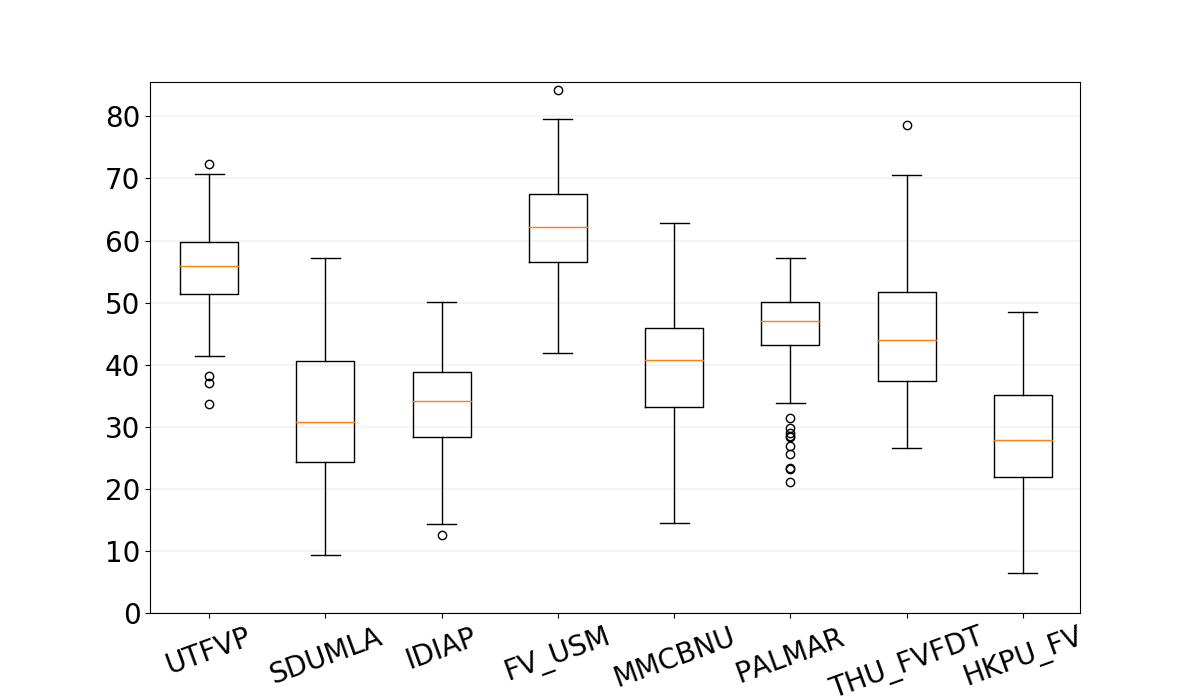}
    \caption{Variance distribution of original and ROI images across all datasets, respectively.}
    
    \label{fig:variance_uncropped}
\end{figure}

To investigate the differences between raw sample data and ROI data in more detail, we have investigated the range of luminance values and their variance across all datasets. Figures \ref{fig: luminance_uncropped} and
\ref{fig:variance_uncropped} display the results in the form of box-plots, where
the left box-plot corresponds to the original raw sample data, and the right one to the ROI data, respectively.
We can clearly see that the luminance distribution properties have been changed dramatically once we change our focus from original datasets to ROI datasets. For example, original HKPU\_FV samples can be discriminated from FV\_USM, MMCBNUm, PALMAR, UTFVP, and THU\_FVFDT ones by just considering luminance value distribution. For the ROI data, the differences are not very pronounced any more. When looking at the variance value distributions, we observe no such strong discrepancy between original sample and ROI data, still for some datasets variance can be used as discrimination criterion (e.g. Palmar vs. HKPU\_FV in original data,
FV\_USM vs. HKPU\_FV in ROI data). Consequently, we expect the discrimination of the considered datasets to be much more challenging when focusing on the ROI data only.

\subsection{\textbf{Pipeline setup and preprocessing}}
\label{ssec:pipline}

Each dataset consists of 120 images, in total 960 images. To enhance the image samples and improve the contrast, we applied Contrast Limited Adaptive Histogram Equalization \textit{(CLAHE)}.
The entire data (960 images from eight datasets) are shuffled and then we take randomly 70\% of data for the training set, 10\% for the evaluation set and 20\% as test set. The splitting policy assures that the data samples used during training is never used during validation or testing. Thus, performance reported is not biased since we have empty intersection among training, validation and test sets. In addition, the \textit{Adam} optimizer is applied for all networks except for Bondi and Marra (as per authors recommendation, the SGD has been applied as an optimizer in both models). Furthermore, batch size is set to be 64. To feed the input to CNN models, we normalized the image samples to uniform width and height, the size of patches for uncropped sample and ROI which are fed to networks is of  $96\times96\times1$. To compare results of CNN-based approaches with a PRNU-based approach, we use the results given in \cite{sollinger2019prnu}. The authors worked on five patches which have been taken from different locations of image samples. For the comparison we consider that results of patch size $320\times240$ are comparable to our results of original image samples. Similarly, the results of patch size $320\times150$ should be comparable to our results of ROI samples.

\subsection{\textbf{Evaluation metrics}}
\label{subsec:evaluation_metrics}
We use classical measures to rate our sensor identification task, which is basically a multi-class classification problem.
We use the area under curve of the receiver operating characteristic ($AUC-ROC$) which relates the false positive rate (FPR) to the false negative rate (FNR). The Analysis of the AUC-ROC is significant as the AUC-ROC shows the ability of the proposed classifier to distinguish classes.

In sensor identification, Precision is a further important measurement metric because it indicates the proportions of positives and negatives, and good result can be interpreted as high performance of a classifier. Also, in the field of biometrics, it is vital to verify the correct sensors (i.e. True Positive). In contrast, it would be a catastrophe if the biometric system verifies the wrong sensor (False Positive). Therefore, Precision is more important than Recall and consequently also used to assess our results.



\section{\textbf{RESULTS}}
\label{sec:result}

In this section, we discuss the result of applying six CNN models on sensor identification of original samples (uncropped datasets) as well as cropped samples (ROI data).

\subsection{\textbf{Results of the six CNN models}}
\label{ssec:results-of-the-six-state-of-the-art-CNN-models}
In the following paragraphs, we will analyze the outcomes of the six mentioned CNN models.  
The first and the second columns of Table \ref{tbl:auc-roc-precision-of-all-models-datasamples}\footnote{Results are rounded to five digits after the decimal point} exhibit the AUC-ROC score of the six applied CNN models on original samples and their corresponding ROI.

\begin{table}[!htbp]
\renewcommand\arraystretch{1.5}
    \begin{center}
    \resizebox{0.50\textwidth}{!}{%
    \begin{tabular}{ccccc}
   
    \multicolumn{1}{l}{\textbf{}}   & \multicolumn{2}{c}{\textbf{AUC-ROC}} &    \multicolumn{2}{c}{\textbf{Precision}} \\
    \cline{2-5}  
    \multicolumn{1}{c}{\textbf{}} &
      \multicolumn{1}{c}{\textbf{Uncropped}} &
      \multicolumn{1}{c}{\textbf{ROI}} &
      \multicolumn{1}{c}{\textbf{Uncropped}} &
      \multicolumn{1}{c}{\textbf{ROI}} \\ \hline  \hline 
    
    \multicolumn{1}{c}{\textit{\textbf{Bondi}}}      & 0.99997 & 0.99773  & 0.9896  & 0.9873 \\ 
    \multicolumn{1}{c}{\textit{\textbf{Marra}}}      & 1.00000 & 0.99856  & 1.0    & 0.9914 \\ 
    \multicolumn{1}{c}{\textit{\textbf{VGG16}}}      & 1.00000 & 0.99945  & 1.0    & 0.9964 \\ 
    \multicolumn{1}{c}{\textit{\textbf{ResNet50}}}   & 0.99996 & 0.99949  & 0.9948 & 0.9971 \\ 
    \multicolumn{1}{c}{\textit{\textbf{Xception}}}   & 1.00000 & 0.99972  & 1.0    & 0.9982 \\ 
    \multicolumn{1}{c}{\textit{\textbf{FV2021}}}     & 1.00000 & 0.99970  & 1.0    & 0.9980  \\ 
    
 \hline\hline
    
    \end{tabular}%
    }
    \caption{\small Results of applied CNN models on ROI and Original (uncropped) samples}
    \label{tbl:auc-roc-precision-of-all-models-datasamples}
    \end{center}
\end{table}%

As was expected, the achieved AUC-ROC scores on original samples are excellent (the first column). All CNN models demonstrated perfect results. However, the modified Bondi model and ResNet(50-layer) results are slightly lower than 1.00 with a small and narrow gap (0.0001). We have the same situation for ROI datasets (the second column). Almost all models exhibited excellent results. Respectively Xception, FV2021, ResNet50 and VGG16 scores are $ > 0.999$. However, concerning the first four mentioned models in the Table, the Marra model and Bondi model results are inferior.

\begin{table}[!htbp]
\renewcommand\arraystretch{1.5}
    \begin{center}
    \resizebox{0.50\textwidth}{!}{%
    \begin{tabular}{llllll}
\textbf{} &
  \textbf{\begin{tabular}[c]{@{}l@{}}PRNU\\  NCC\end{tabular}} &
  \textbf{\begin{tabular}[c]{@{}l@{}}PRNU\\  PCE\end{tabular}} &
  \textbf{\begin{tabular}[c]{@{}l@{}}Texture\\  Descriptor\\  (WMV)\end{tabular}} &
  \textbf{\begin{tabular}[c]{@{}l@{}}Deep \\ Learning\\ (Xception)\end{tabular}} &
  \textbf{\begin{tabular}[c]{@{}l@{}}Deep\\ Learning\\ (FV2021)\end{tabular}} \\
  \hline\hline
  
\textbf{\begin{tabular}[c]{@{}l@{}} Original \\ Sample\end{tabular}} &
  0.992 &  0.991 &  0.999 &  1.0 &  1.0 \\
  
\textbf{} &   &   &   &   &  \\
  
\textbf{ROI} &
  0.998 &  0.997 &  0.994 &  0.99972 &  0.99970 \\
\hline\hline
    \end{tabular}
    }
    \caption{\small Comparing results of PRNU, Texture Descriptor, and Deep Learning (CNN) methods on original and ROI sample data.}
    \label{tbl:prnu_texture_descript_cnn}
    \end{center}
\end{table}

The third column of Table \ref{tbl:auc-roc-precision-of-all-models-datasamples} displays the Precision score of all CNN models on original samples (uncropped datasets). Respectively Xception model, FV2021 model, VGG16 model and Marra model are superior to ResNet50 and Bondi's model. The Precision scores are either $1.0$ or close to $1.0$. As a result, by observing Precision results, we can imply that obtained results by four models are highly reliable and accurate.

Moving from the original samples (uncropped) to the ROI, the fourth column of the Table \ref{tbl:auc-roc-precision-of-all-models-datasamples} exhibits the performance of the applied models on the region of interest. By observing Precision scores, respectively the Precision score of Xception model $\simeq$ FV2021 model $>$ ResNet50 model $>$ VGG16 model $>$ Marra model $>$ Bondi model. Thus the Xception model and the proposed FV2021 model improved results slightly as compared to the others. 

We would like to emphasize that the value of false positives (FP) in ResNet50, VGG16, Marra and Bondi models are relatively high which causes their Precision scores to get lower than this of Xception and FV2021 models.

\subsection{\textbf{Comparison of various approaches}}
\label{ssec:comparing_prnu_texture_descript_and_cnn}

In this section, we compare the performance of various approaches for identifying the FV image origin. As we explained in Section \ref{ssec:pipline}, we compare to the results of a PRNU-based approach which is proposed by Söllinger \etal \cite{sollinger2019prnu} and a texture-based approach which is done by Maser \etal \cite{maser2021identifying}.
Table \ref{tbl:prnu_texture_descript_cnn} shows results of these three approaches.  We observe the superiority of deep learning (CNN) methods using the proposed FV2021 and the Xception CNN models, respectively, over the PRNU-based approach and the texture-based approach. FV2021 and Xception models compete closely in the race, their results are approximately equal, but due to the significantly lower complexity of FV2021 (which has been approved by analysing the number of trainable parameters in Table \ref{tbl:number-of-params-in-cnn-model}), we take the FV2021 as the superior CNN model.  


\subsection{\textbf{Single Sensor-based Results}}
\label{ssec:Sensor-based-results}
In this section, Table \ref{tbl:results-sensor-based-Uncropped} displays results of the employed CNN models for all uncropped datasets (instead of overall results shown before).
All sensors are discriminated ideally except than MMCBNU. The ResNet50 and Bondi experienced some difficulties to discriminate the MMCBNU sensor.\\  
We observe results of the employed CNN models for all ROI datasets in Table \ref{tbl:results-sensor-based-roi}. The excellent performance of Xception model and FV2021, ResNet50 on all sensors can be seen. among these results only FV2021 was successful to gain the results either at 1.0 or at 0.999 on every sensor. 

\begin{table}[!htbp]
\renewcommand\arraystretch{1.5}
    \begin{center}
    \resizebox{0.50\textwidth}{!}{%
    \begin{tabular}{l||llllll}
    \textbf{Sensor} & \textbf{Bondi} & \textbf{Marra} & \textbf{VGG16} & \textbf{ResNet50} & \textbf{Xception} & \textbf{FV2021} \\
    \hline \hline
\textbf{UTFVP}      & 1.0000         & 1.0             &  1.0          & 1.0               & 1.00               & 1.0             \\
\textbf{FV\_USM}    & 1.0000         & 1.0             &  1.0          & 1.0               & 1.00               & 1.0             \\
\textbf{PALMAR}     & 1.0000         & 1.0             &  1.0          & 1.0               & 1.00               & 1.0             \\
\textbf{SDUMLA}     & 1.0000         & 1.0             &  1.0          & 1.0               & 1.00               & 1.0             \\
\textbf{THU\_FVFDT} & 1.0000         & 1.0             &  1.0          & 1.0               & 1.00               & 1.0             \\
\textbf{IDIAP}      & 1.0000         & 1.0             &  1.0          & 1.0               & 1.00               & 1.0             \\
\textbf{MMCBNU}     & 0.9997         & 1.0             &  1.0          & 1.0               & 0.96               & 1.0             \\
\textbf{HKPU-FV}    & 1.0000         & 1.0             &  1.0          & 1.0               & 1.00               & 1.0             \\

    \end{tabular}
    }
    \caption{ Displaying results (AUC-ROC score) of applied CNN models on all uncropped FV databases.}
    \label{tbl:results-sensor-based-Uncropped}
 
\end{center}
\end{table}

\begin{table}[!htbp]
\renewcommand\arraystretch{1.5}
    \begin{center}
    \resizebox{0.50\textwidth}{!}{%
    \begin{tabular}{l||llllll}
        \textbf{Sensors} &   \textbf{Bondi}& \textbf{Marra}& \textbf{VGG16} & \textbf{ResNet50}   & \textbf{Xception}& \textbf{FV2021}  \\
        \hline \hline
\textbf{UTFVP}           & 0.9885          & 0.9985        & 0.9987         & 0.9997              & 0.9985           & 0.9992           \\
\textbf{FV\_USM}         & 0.9997          & 0.9997        & 1.0000         & 1.0000              & 1.0000           & 1.0000           \\
\textbf{PALMAR}          & 1.0000          & 0.9997        & 1.0000         & 1.0000              & 1.0000           & 1.0000           \\
\textbf{SDUMLA}          & 0.9992          & 0.9994        & 1.0000         & 1.0000              & 0.9994           & 1.0000           \\
\textbf{THU\_FVFDT}      & 0.9991          & 0.9964        & 0.9997         & 0.9997              & 1.0000           & 0.9993           \\
\textbf{IDIAP}           & 0.9975          & 0.9987        & 0.9997         & 0.9982              & 1.0000           & 0.9990           \\
\textbf{MMCBNU}          & 0.9995          & 1.0000        & 0.9976         & 0.9997              & 1.0000           & 0.9997           \\
\textbf{HKPU-FV}         & 0.9980          & 0.9990        & 0.9992         & 1.0000              & 1.0000           & 1.0000           \\
\end{tabular}
}
 \caption{ Displaying results (AUC-ROC score) of applied CNN models on all ROI FV databases.}
     \label{tbl:results-sensor-based-roi}
\end{center}
\end{table}

\section{\textbf{Conclusion}}
\label{sec:conclution}

In this research, we studied the results of using five $state-of-the-art$ CNN models and a novel CNN model (FV2021) for sensor identification on the ROI as well as the original finger vein samples. Finger vein samples are taken from eight databases. As a result, the performance of the proposed FV2021 and Xception models are superior to other CNN models. Then we compare CNN-based results with other  results including PRNU correlation-based and texture descriptor-based research. The CNN-based results show slightly better performance. Besides, the two top performing CNN architectures perform very closely in terms of sensor identification accuracy but due to much lower model complexity, we recommend the proposed FV2021. The achieved result by FV2021 is excellent, i.e., the AUC-ROC score for ROI data is 0.9997 and for original samples it is at 1.0.


{\small
\bibliography{egbib}

\begin{thebibliography}{10}

\bibitem{maser2021identifying}
Babak Maser and Andreas Uhl.
\newblock Identifying the origin of finger vein samples using texture
  descriptors.
\newblock {\em arXiv preprint arXiv:2102.03992}, 2021.

\bibitem{lukas2006digital}
Jan Lukas, Jessica Fridrich, and Miroslav Goljan.
\newblock Digital camera identification from sensor pattern noise.
\newblock {\em IEEE Transactions on Information Forensics and Security},
  1(2):205--214, 2006.

\bibitem{fridrich2009digital}
Jessica Fridrich.
\newblock Digital image forensics.
\newblock {\em IEEE Signal Processing Magazine}, 26(2):26--37, 2009.

\bibitem{ahmed2019comparative}
Farah Ahmed, Fouad Khelifi, Ashref Lawgaly, and Ahmed Bouridane.
\newblock Comparative analysis of a deep convolutional neural network for
  source camera identification.
\newblock In {\em 2019 IEEE 12th International Conference on Global Security,
  Safety and Sustainability (ICGS3)}, pages 1--6. IEEE, 2019.

\bibitem{baroffio2016camera}
Luca Baroffio, Luca Bondi, Paolo Bestagini, and Stefano Tubaro.
\newblock Camera identification with deep convolutional networks.
\newblock {\em arXiv preprint arXiv:1603.01068}, 2016.

\bibitem{tuama2016camera}
Amel Tuama, Fr{\'e}d{\'e}ric Comby, and Marc Chaumont.
\newblock Camera model identification with the use of deep convolutional neural
  networks.
\newblock In {\em 2016 IEEE International workshop on information forensics and
  security (WIFS)}, pages 1--6. IEEE, 2016.

\bibitem{bondi2016first}
Luca Bondi, Luca Baroffio, David G{\"u}era, Paolo Bestagini, Edward~J Delp, and
  Stefano Tubaro.
\newblock First steps toward camera model identification with convolutional
  neural networks.
\newblock {\em IEEE Signal Processing Letters}, 24(3):259--263, 2016.

\bibitem{5204312}
N.~{Bartlow}, N.~{Kalka}, B.~{Cukic}, and A.~{Ross}.
\newblock Identifying sensors from fingerprint images.
\newblock In {\em 2009 IEEE Computer Society Conference on Computer Vision and
  Pattern Recognition Workshops}, pages 78--84, 2009.

\bibitem{Kauba18c}
Christof Kauba, Bernhard Prommegger, and Andreas Uhl.
\newblock Focussing the beam - a new laser illumination based data set
  providing insights to finger-vein recognition.
\newblock In {\em Proceedings of the IEEE 9th International Conference on
  Biometrics: Theory, Applications, and Systems (BTAS2018)}, pages 1--9, Los
  Angeles, California, USA, 2018.

\bibitem{banerjee2017image}
Sudipta Banerjee and Arun Ross.
\newblock From image to sensor: Comparative evaluation of multiple prnu
  estimation schemes for identifying sensors from nir iris images.
\newblock In {\em 2017 5th International Workshop on Biometrics and Forensics
  (IWBF)}, pages 1--6. IEEE, 2017.

\bibitem{marra2018deep}
Francesco Marra, Giovanni Poggi, Carlo Sansone, and Luisa Verdoliva.
\newblock A deep learning approach for iris sensor model identification.
\newblock {\em Pattern Recognition Letters}, 113:46--53, 2018.

\bibitem{maser2019prnu}
Babak Maser, Dominik S{\"o}llinger, and Andreas Uhl.
\newblock Prnu-based detection of finger vein presentation attacks.
\newblock In {\em 2019 7th International Workshop on Biometrics and Forensics
  (IWBF)}, pages 1--6. IEEE, 2019.

\bibitem{maser2019prnuOAGM}
Babak Maser, Dominik S{\"o}llinger, and Andreas Uhl.
\newblock Prnu-based finger vein sensor identification in the presence of
  presentation attack data.
\newblock In {\em Proceedings of the Joint ARW/OAGM Workshop}, volume 2019,
  2019.

\bibitem{sollinger2019prnu}
Dominik S{\"o}llinger, Babak Maser, and Andreas Uhl.
\newblock Prnu-based finger vein sensor identification: On the effect of
  different sensor croppings.
\newblock In {\em 2019 International Conference on Biometrics (ICB)}, pages
  1--8. IEEE, 2019.

\bibitem{goodfellow2013multi}
Ian~J Goodfellow, Yaroslav Bulatov, Julian Ibarz, Sacha Arnoud, and Vinay Shet.
\newblock Multi-digit number recognition from street view imagery using deep
  convolutional neural networks.
\newblock {\em arXiv preprint arXiv:1312.6082}, 2013.

\bibitem{simonyan2014very}
Karen Simonyan and Andrew Zisserman.
\newblock Very deep convolutional networks for large-scale image recognition.
\newblock {\em arXiv preprint arXiv:1409.1556}, 2014.

\bibitem{he2016deep}
Kaiming He, Xiangyu Zhang, Shaoqing Ren, and Jian Sun.
\newblock Deep residual learning for image recognition.
\newblock In {\em Proceedings of the IEEE conference on computer vision and
  pattern recognition}, pages 770--778, 2016.

\bibitem{chollet2017xception}
Fran{\c{c}}ois Chollet.
\newblock Xception: Deep learning with depthwise separable convolutions.
\newblock In {\em Proceedings of the IEEE conference on computer vision and
  pattern recognition}, pages 1251--1258, 2017.

\end{thebibliography}

\bibliographystyle{unsrt}
}

\end{document}